\documentclass[10pt]{article}
\usepackage[top=1in, bottom=1in, left=1in, right=1in]{geometry}
\usepackage{amssymb,amsfonts,amsmath,amsthm,amscd,dsfont,mathrsfs,mathtools,microtype,nicefrac,pifont,natbib,soul}
\usepackage[linktocpage,colorlinks,linkcolor=blue,anchorcolor=blue,citecolor=blue,urlcolor=blue,pagebackref]{hyperref}
\renewcommand*{\backref}[1]{\ifx#1\relax \else Page #1 \fi}
\renewcommand*{\backrefalt}[4]{%
  \ifcase #1 \footnotesize{(Not cited.)}%
  \or        \footnotesize{(Cited on page~#2.)}%
  \else      \footnotesize{(Cited on pages~#2.)}%
  \fi
}
\usepackage[utf8]{inputenc} 
\usepackage[T1]{fontenc}    
\usepackage{hyperref}       
\usepackage{url}            
\usepackage{booktabs}       
\usepackage{amsfonts}       
\usepackage{nicefrac}       
\usepackage{microtype}      
\usepackage{xcolor}         
\usepackage{authblk}
\usepackage{algorithm}
\usepackage{algpseudocode}
\usepackage{bbm}
\usepackage{bm}
\usepackage{color, colortbl}
\definecolor{LightCyan}{rgb}{0.88,1,1}
\usepackage{dirtytalk}
\usepackage{enumerate}
\usepackage{graphicx}
\usepackage{listings}
\usepackage{subfigure}
\usepackage{enumitem}
\usepackage{xspace}
\usepackage{makecell}
\usepackage{multirow, array}
\usepackage{verbatim}
\usepackage[capitalize,noabbrev]{cleveref}
\usepackage{tikz}
\usepackage{tcolorbox}
\usepackage{tablefootnote}
\usepackage{cancel}
\usepackage[symbol]{footmisc}

\usepackage{adjustbox}
\usepackage{caption}
\usepackage{wrapfig}

\newtheorem{lemma}{Lemma}[section]

\theoremstyle{remark}

\newcommand{\cA}{\mathcal{A}}
\newcommand{\cB}{\mathcal{B}}

\newcommand{\cO}{\mathcal{O}}

\newcommand{\E}{\mathbb{E}}

\newcommand{\norm}[1]{\left\|#1\right\|}

\newcommand{\T}{^\top}

\newtcolorbox{thmbox}{colback=cyan!5,colframe=white}
\newtcolorbox{questionbox}{colback=red!5!white,colframe=white}
\newtcolorbox{updatebox}{colback=white,colframe=black}

\DeclareMathOperator*{\argmin}{arg\,min\,}

\def\<#1,#2>{\left\langle #1,#2\right\rangle}
\mathchardef\mhyphen="2D
\allowdisplaybreaks

\newcommand{\B}{\operatorname{\mathbb B}}
\newcommand{\R}{\operatorname{\mathbb R}}

\newcommand{\bS}{\operatorname{\mathbb S}}

\newcommand{\softmax}{\operatorname{softmax}}

\newcommand{\range}{\operatorname{range}}

\DeclarePairedDelimiterX{\infdivx}[2]{(}{)}{%
  #1\;\delimsize\|\;#2%
}

\theoremstyle{plain}        
\newtheorem{thm}{Theorem}[section]

\theoremstyle{definition}     
\newtheorem{defn}[thm]{Definition}

\title{\bf{Feed $m$ Birds with One Scone: Accelerating Multi-task Gradient Balancing via Bi-level Optimization}\footnote{Work done when the first author was doing an internship at Meta in 2024 spring.}}

\author{Xuxing Chen\thanks{Corresponding author. Email: \texttt{xxchen3636@gmail.com}}}
\author{Yun He}
\author{Jiayi Xu}
\author{Minhui Huang}
\author{Xiaoyi Liu}
\author{Boyang Liu}
\author{Fei Tian}
\author{Xiaohan Wei}
\author{Rong Jin}
\author{Sem Park}
\author{Bo Long}
\author{Xue Feng}
\affil{Meta}

\begin{document}
\date{ }
\maketitle

\begin{abstract}
    In machine learning, the goal of multi-task learning (MTL) is to optimize multiple objectives together. Recent works, for example, Multiple Gradient Descent Algorithm (\texttt{MGDA}) and its variants, show promising results with dynamically adjusted weights for different tasks to mitigate conflicts that may potentially degrade the performance on certain tasks. Despite the empirical success of \texttt{MGDA}-type methods, one major limitation of such methods is their computational inefficiency, as they require access to all task gradients. In this paper we introduce \texttt{MARIGOLD}, a unified algorithmic framework for efficiently solving MTL problems. Our method reveals that multi-task gradient balancing methods have a hierarchical structure, in which the model training and the gradient balancing are coupled during the whole optimization process and can be viewed as a bi-level optimization problem.
    Moreover, we showcase that the bi-level problem can be solved efficiently by leveraging zeroth-order method. Extensive experiments on both public datasets and industrial-scale datasets demonstrate the efficiency and superiority of our method.
\end{abstract}

\section{Introduction}
The study of multi-task learning (MTL) can be traced back to~\citep{caruana1997multitask}. Instead of conducting training based on one single task, MTL aims at optimizing different loss functions simultaneously in one training algorithm, and it has been widely used in many areas like natural language processing~\citep{mccann2018natural}, computer vision~\citep{liu2019end}, recommendation systems~\citep{wang2018explainable}, reinforcement learning~\citep{yu2020gradient}, etc. Through the lens of multi-objective optimization (MOO)~\citep{miettinen1999nonlinear, sener2018multi}, one may aim at solving the following problem.
\begin{align}\label{opt: MOO}
    \min_{\theta\in \R^d} F(\theta) := \left(f_1(\theta), ..., f_m(\theta)\right)^\top
\end{align}
where we denote by $f_i$ the loss function for task $i$, $m$ the number of tasks, and $\theta$ the trainable model parameters. It is often impossible to find one $\theta$ that minimizes all $f_i$, thus a practical way is to linearly combine all loss functions via a weight vector $\lambda$, which gives the following problem.
\begin{align}\label{opt: MTL_linear}
    \min_{\theta\in \R^d}\ \sum_{i=1}^{m}\lambda_if_i(\theta).
\end{align}
It is believed that multiple similar tasks may share a common learnable representation parameterized by deep neural networks, and thus optimizing the aggregation of multiple tasks can potentially enable better representation learning, which has also been theoretically investigated in certain scenarios~\citep{du2020few, zhang2023sample, collins2023provable}.

In the deep learning era, MTL is gaining more and more popularity thanks to its capability of efficiently handling multiple tasks simultaneously with superior generalization performance. While we witness the great success of MTL, one major challenge in designing MTL algorithms lies at the conflicts of gradients between different tasks, i.e., $\<\nabla f_i(\theta), \nabla f_j(\theta)> < 0$ for some $i,j$. This type of conflicts can potentially lead to negative transfer~\citep{li2023identification}. For example, updating the model parameters via gradient descent merely along $-\nabla f_i$ or $-\nabla f_j$ may increase the other loss. It is thus critical to carefully design the task weights that can well reflect the intricate trade-offs between different tasks.

Based on different information utilized during the training process, multi-task balancing can be roughly divided into two categories -- loss balancing and gradient balancing. Loss balancing methods keep track of all function values $f_i$ and derives the task weights based on different criteria, such as uncertainty~\citep{kendall2018multi}, rate of change~\citep{liu2019end}, etc. In comparison, gradient balancing methods further require the information of different task gradients so that one can design the weights, or manipulate the gradients based on different criteria, such as gradient manipulation~\citep{chen2018gradnorm, yu2020gradient, chen2020just, wang2021gradient, he2022metabalance}, conflict-averse direction~\citep{liu2021conflict}, game theory~\citep{navon2022multi}, etc.

It has been shown that loss balancing methods are often more efficient than gradient balancing ones in terms of time and space complexity, as the former only requires the loss function values for balancing while the latter further requires computing and storing gradients of all tasks, which lead to $\cO(md)$ time and space complexity. Despite the inefficiency, gradient balancing methods have better performance both empirically~\citep{liu2021conflict, navon2022multi} and theoretically~\citep{sener2018multi, zhou2022convergence, fernando2022mitigating, chen2023three, xiao2023direction}. In this paper, we focus on improving the time and space complexity of gradient balancing in MTL, while maintaining its superiority over loss balancing.
In particular, our contributions can be summarized as follows.
\begin{itemize}[leftmargin=*]
    \item We propose \texttt{MARIGOLD}, a unified multi-task gradient balancing framework via bi-level optimization, which greatly reduces the per-iteration time and space complexity from $\cO(m d)$ in most gradient balancing techniques to $\cO(d)$. Our algorithm is also model-agnostic, in the sense that it is compatible with any MTL models trained with any optimizers.
    \item We conduct extensive experiments on both public and industrial-scale datasets, which empirically validate the superiority of our algorithms in terms of performance and efficiency.
\end{itemize}

\section{Preliminaries}
Before we present the main building blocks of our algorithms, we first provide some preliminaries on MTL, MOO and bi-level optimization.


\begin{figure}[t]
    \centering
    \subfigure[MTL architecture]{\includegraphics[width=0.55\textwidth]{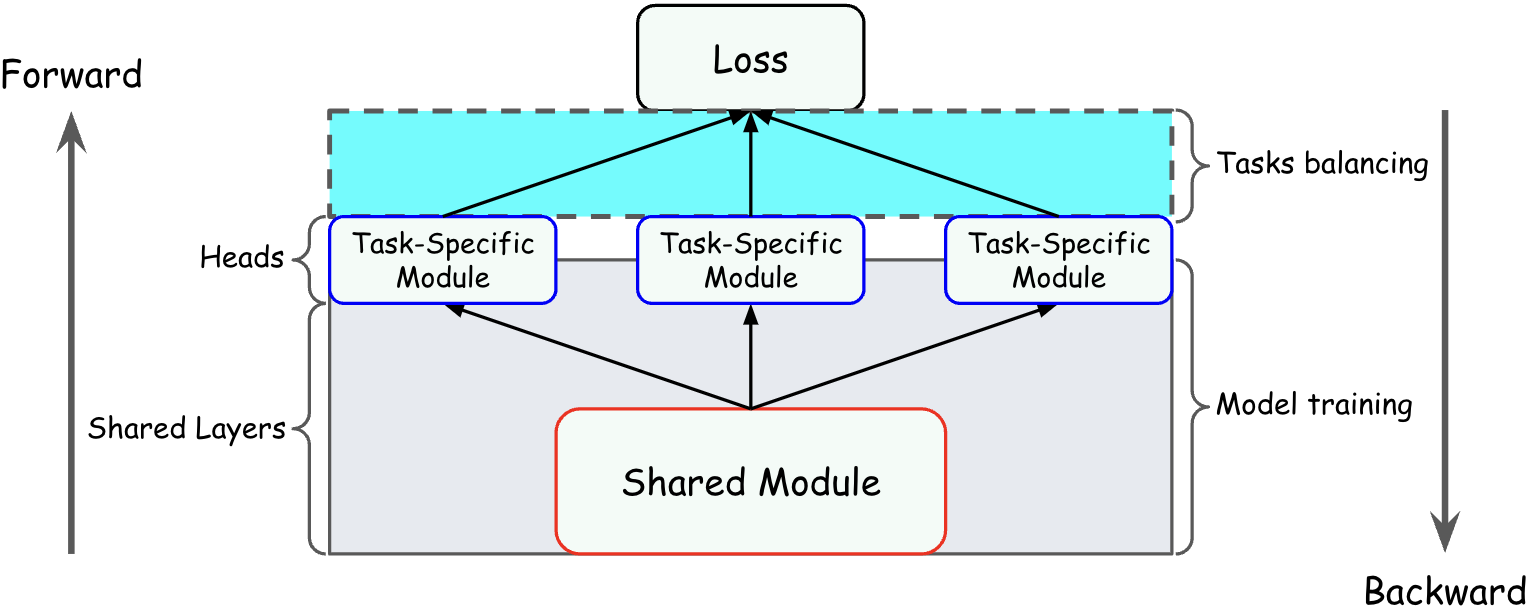}\label{fig: arch}}
    \subfigure[\texttt{MGDA}-type algorithms]{\includegraphics[width=0.44\textwidth]{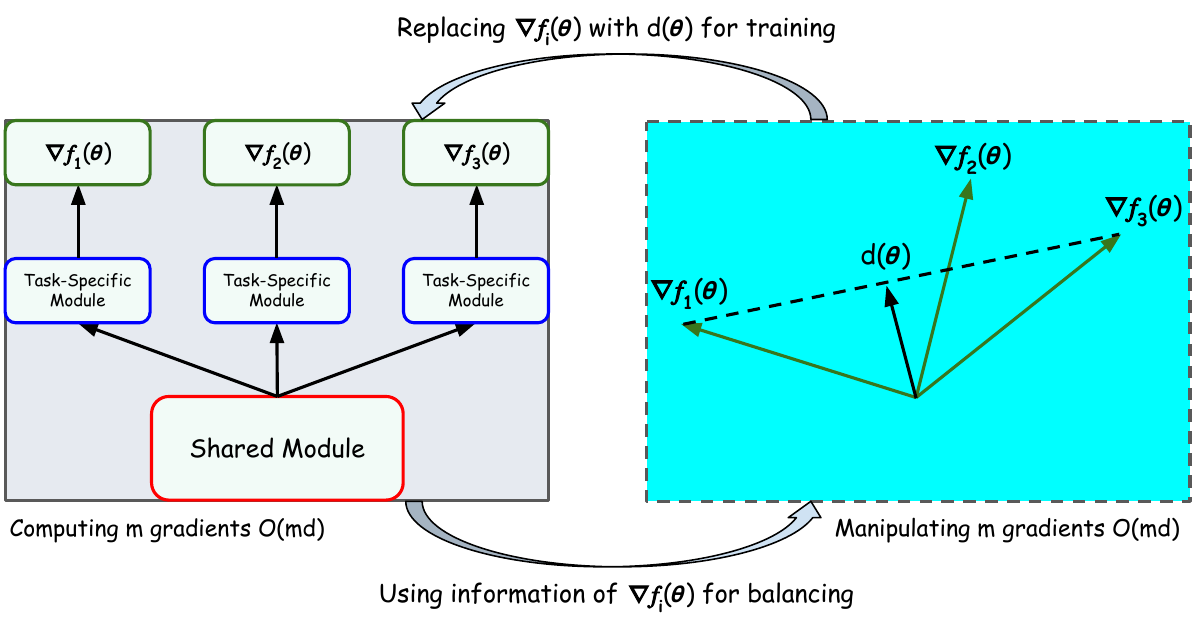}\label{fig: mgda}}
    \caption{In Figure \ref{fig: arch} we have a sketch of a model architecture for multi-task learning. In the forward pass, a batch of data is fed into the bottom layers shared by all tasks to provide a representation for task-specific modules (also known as ``heads''). Then task-specific losses are constructed based on the prediction output of heads. Finally a linear combination of all losses gives the main objective function to optimize. In the backward pass, model parameters receive gradients and are updated via certain optimization algorithms as the model training step, and the task weights are balanced based on pre-specified criterion as the tasks balancing step. In Figure \ref{fig: mgda} we showcase how \texttt{MGDA}-type algorithms~\citep{sener2018multi} work -- the gradient balancing step tries to find a convex combination of gradients with minimum norm, and updates the model parameters accordingly.}
    \label{fig: mtl_arch}
    \vspace{-0.2cm}
\end{figure}

\textbf{Notation.}\quad In this paper, we denote by $\Delta^m:=\{\lambda\in\R^m: \sum_{i=1}^m\lambda_i=1, \lambda_i\geq 0\}$ the probability simplex in $\R^m$. We use $\norm{\cdot}$ to denote $\ell^2$-norm for vectors. For a matrix $A\in \R^{p\times q}$ we write $\range(A) = \{Ax: x\in\R^q\}$ as the range of the linear mapping induced by matrix $A$. For a loss function $f_i(\theta)$ and a batch of data $\cB$, we use $f_i(\theta;\cB)$ and $\nabla f_i(\theta;\cB)$ to represent the function value and gradient evaluated on $\cB$.

\subsection{Multi-task learning and multi-objective optimization}
Consider the multi-objective optimization problem in \eqref{opt: MOO}, we have the following notion of optimality and stationarity, which are standard in MOO literature~\citep{desideri2012multiple, sener2018multi, zhou2022convergence, fernando2022mitigating}.
\begin{defn}
    For $m$ given objective functions $f_i(\theta)$ in \eqref{opt: MOO}, we say a point $\theta\in \R^d$ is {\bf Pareto optimal}, when there is no $\theta'\in \R^d$ such that $f_i(\theta')\leq f_i(\theta)$ for any $1\leq i \leq m$ and $F(\theta)\neq F(\theta')$. A point $\theta\in \R^d$ is called {\bf Pareto stationary}, when $\range(\nabla F(\theta)\T) \bigcap (-\R_{++}^m)$ is an empty set. In other words, there does not exist a vector $v\in \R^d$, such that every coordinate of $\nabla F(\theta)\T v$ is negative. 
\end{defn}

Typically, an MTL algorithm aims at finding the Pareto stationary point of Problem \eqref{opt: MOO} via solving \eqref{opt: MTL_linear}, in which the task weights $\lambda = (\lambda_1, ..., \lambda_m)^\top$ are chosen via a carefully-designed subroutine. The model architecture is usually designed to first learn a common pattern across different tasks via a shared module, whose output is then fed into task-specific modules that produce different losses defined in \eqref{opt: MOO}. Finally a linear combination of all losses in \eqref{opt: MTL_linear} is constructed as the main objective function to optimize. A sketch of the whole MTL paradigm is depicted in Figure \ref{fig: mtl_arch}.

There has been a flurry of work presenting promising results of various task weights balancing techniques like loss balancing~\citep{liu2019end, kendall2018multi, liu2021towards} and gradient balancing~\citep{desideri2012multiple, yu2020gradient, chen2020just, liu2021conflict, liu2021towards, zhou2022convergence, fernando2022mitigating, chen2023three, navon2022multi, liu2024famo, xiao2023direction}. Theoretically, deep neural networks usually lead to highly non-convex optimization problems, which makes it impossible to find Pareto optimal points. Thus we may only aim at finding Pareto stationary points in deep learning applications. Under certain assumptions, it can be shown that \texttt{MGDA}-type algorithms can provably find Pareto stationary points~\citep{liu2021stochastic, zhou2022convergence, fernando2022mitigating, chen2023three, xiao2023direction}. In addition to the theoretical results, existing works have witnessed the empirical success of gradient balancing methods~\citep{sener2018multi, navon2022multi, liu2024famo, xiao2023direction} over others. However, one major concern that slows down the training of MTL problems is that, gradient balancing typically requires intricate manipulations over all task gradients, which incurs $\cO(md)$ time and space complexity, which motivates some followup works~\citep{sener2018multi, liu2024famo} to mitigate this issue. We defer a detailed discussion on this to Section~\ref{sec: grad_balance}.


\begin{figure}
    \centering
    \includegraphics[width=0.7\textwidth]{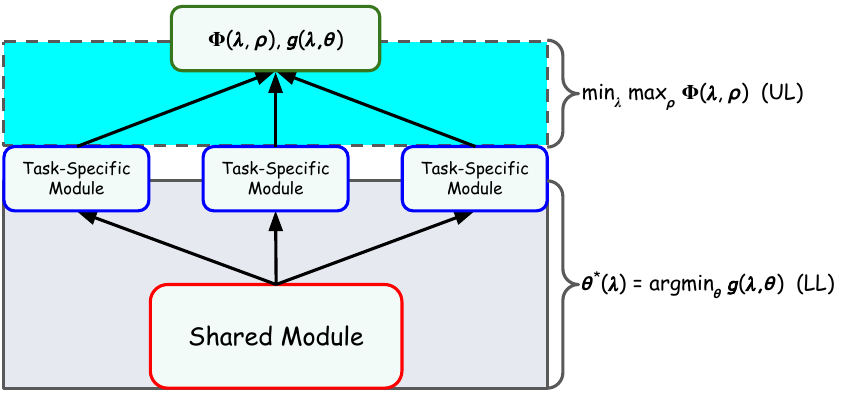}
    \caption{MTL as a bi-level optimization problem. The Upper-Level (UL) function $\Phi(\lambda, \rho)$ and Lower-Level (LL) function $g(\lambda,\rho)$ are defined in \eqref{opt: bo_ul} and \eqref{opt: bo_ll}.}
    \label{fig: bo4mtl}
    \vspace{-0.2cm}
\end{figure}

\subsection{Bi-level optimization}
Problems like meta learning, hyperparameter optimization and reinforcement learning sometimes cannot be formulated as a single-level optimization problem, as it fails to capture the intricacy of the interacting parameters coming from different sources. Taking gradient-based hyperparameter optimization~\citep{bengio2000gradient} as an example, we may see that hyperparameters should be tuned on validation loss while model parameters are optimized over the training loss.

Motivated by the limitation of single-level optimization, both optimization and machine learning community have proven that bi-level optimization successfully handles the nested structure in these problems~\citep{rajeswaran2019meta, lorraine2020optimizing, hong2023two}. Typically, bi-level optimization problems can be written as follows.
\begin{align*}
    &\min_{x}\  \phi(x) := f_{\text{upper}}(x, y^*(x)),\quad \text{s.t. }\ y^*(x) := \argmin_{y} f_{\text{lower}}(x, y),
\end{align*}
in which we denote by $f_{\text{upper}}, f_{\text{lower}}$ the upper-level and lower-level objective function respectively. Assuming $f_{\text{lower}}$ is strongly-convex in $y$, one may design algorithms that can provably find stationary points of the above problem in finite time~\citep{ghadimi2018approximation, hong2023two, chen2021closing, dagreou2022framework, chen2023optimal, hao2023bilevel}.

\section{Methodology}\label{sec: method}

In this section we introduce the basic ideas of our proposed framework \texttt{MARIGOLD}. We first review basic criteria for gradient balancing in Section \ref{sec: mtl_criterion}, and then reveal the intrinsic bi-level structure on top of it in Section \ref{sec: mtl_bo}. The main algorithms are then presented in Sections \ref{sec:hypergrad_zo_grad} and \ref{sec: algorithm}. We finally describe a generalized version of our framework that can handle different variants of MTL problems.
  
\subsection{Rethinking gradient balancing criterion}\label{sec: mtl_criterion}
 We first make an observation that most of the ideas of gradient balancing methods in MTL, such as \texttt{MGDA}-type methods~\citep{sener2018multi, zhou2022convergence, fernando2022mitigating, xiao2023direction} and gradient manipulation based methods~\citep{yu2020gradient, chen2020just, liu2021conflict, liu2021towards, navon2022multi}, start with manipulating gradients across different tasks. We take \texttt{CAGrad}~\citep{liu2021conflict} as an illustrative example, which seeks to minimize the worst-case decrement among all losses. Suppose we have an underlying training algorithm $\cA$, which takes the model parameter $\theta$ and task weights $\lambda$ as input and outputs the updated model parameter. For example, we may choose $\cA$ to be one-step gradient descent for solving \eqref{opt: MTL_linear}, i.e.,
\begin{align}\label{eq: A_sgd}
    \cA(\lambda, \theta) = \theta - \alpha \sum_{i=1}^{m}\lambda_i\nabla f_i(\theta).
\end{align}
We further define the worst-case decrement, which can be seen as a generalization of that in \texttt{CAGrad}~\cite{liu2021conflict}:
\begin{align}\label{eq: worst_case_decrement}
    R(\lambda, \theta) := \max_{1\leq i\leq m}(f_i(\cA(\lambda, \theta)) - f_i(\theta)) = \max_{\rho\in \Delta^m}\sum_{i=1}^{m}\rho_i(f_i(\cA(\lambda, \theta)) - f_i(\theta)),
\end{align}
where the second equality holds since $\rho\in \Delta^m$. Following~\cite{liu2021conflict}, we aim at solving for $\lambda$ such that the decrement $R(\lambda, \theta)$ is minimized.
\begin{align}
    \min_{\lambda\in \Delta^m} R(\lambda, \theta) = \min_{\lambda\in \Delta^m}\max_{\rho\in \Delta^m}\sum_{i=1}^{m}\rho_i(f_i(\cA(\lambda, \theta)) - f_i(\theta)). \label{opt: minmax_ul}
\end{align}
In most existing works~\citep{liu2021conflict, zhou2022convergence, xiao2023direction}, by assuming $\cA(\lambda, \theta) = \theta - \alpha d(\lambda, \theta)$, where $d(\lambda, \theta)$ denotes the update direction for $\theta$, one extra linearization step can be performed on $f_i(\cA(\lambda, \theta)) - f_i(\theta)$ as
\begin{align}\label{eq: linearization}
    f_i\left(\theta - \alpha d(\lambda, \theta)\right) - f_i(\theta)\approx -\alpha \<\nabla f_i(\theta), d(\lambda, \theta)>.
\end{align}
It has been shown that under this approximation, to achieve Pareto stationarity, $\cA(\lambda, \theta)$ should be chosen as the one-step gradient descent in \eqref{eq: A_sgd} given certain constraint~\citep{liu2021conflict}. Note that in this case $\cA(\lambda, \theta)$ is linear in $\lambda$, and the linearity simplifies the problem and makes \eqref{opt: minmax_ul} more tractable. Besides, it has been proven~\citep{xiao2023direction} that \texttt{CAGrad} is closely related to \texttt{MGDA}-type methods, and thus is capable of finding Pareto stationary points under standard assumptions in optimization theory.

However, there are several restrictions of this type of method. First, $\alpha$ should be chosen extremely small to control the error incurred by the approximation in \eqref{eq: linearization}, while in certain cases large stepsizes can lead to better generalization performance~\citep{cohen2020gradient}. Second, the optimizer $\cA(\lambda, \theta)$ for model training should be chosen as gradient descent according to the theory (see, e.g., Algorithm 1 in \cite{liu2021conflict} and Algorithm 1 in \cite{xiao2023direction}), which is inconsistent with the actual implementation that uses \texttt{Adam}~\citep{kingma2014adam}.


Here we do not adopt the linearization strategy in \eqref{eq: linearization}. Instead we note that objective function in \eqref{opt: minmax_ul} is non-convex in $\lambda$ and concave in $\rho$. Thus we could apply non-convex-concave min-max optimization algorithms. One possible scheme widely used in optimization literature is alternatingly updating $\lambda$ and $\rho$~\citep{lin2020gradient}. We highlight here that, min-max optimization typically only requires the gradients of \eqref{opt: minmax_ul} with respect to $\lambda$ and $\rho$, and computing the gradient with respect to $\lambda$ in either \eqref{opt: minmax_ul} only requires computing one gradient in each iteration, thanks to the composite structure which combines the interactions between different gradients into a single hypergradient. This greatly improves the per-iteration complexity of most existing gradient balancing methods~\citep{sener2018multi, chen2018gradnorm, liu2019end, yu2020gradient, chen2020just, liu2021stochastic, wang2021gradient, liu2021conflict, navon2022multi, fernando2022mitigating, xiao2023direction, chen2023three}, which typically require $m$ forward-backward passes in each iteration to update $\lambda$. 

Besides, $\cA(\lambda, \theta)$ can be chosen as any optimizer to update the model parameters $\theta$, which provides more flexibility to incorporate it into industrial-scale systems, in which optimizers like \texttt{Adam}~\citep{kingma2014adam} and \texttt{AdaGrad}~\citep{duchi2011adaptive} are widely used.


\subsection{Multi-task gradient balancing as bi-level optimization}\label{sec: mtl_bo}
Now that we have shown the gradient balancing step can be viewed as a min-max optimization problem without requiring any linear approximation steps, we can further combine the model training step with balancing via the following bi-level optimization problem, which is depicted in Figure \ref{fig: bo4mtl}.
\begin{align}
    \min_{\lambda\in \Delta^m}\max_{\rho\in \Delta^m}\ \Phi(\lambda, \rho) &:= \sum_{i=1}^{m}\rho_i(f_i(\cA(\lambda, \theta^*(\lambda))) - f_i(\theta^*(\lambda))) \qquad &\text{Upper Level (UL)}, \label{opt: bo_ul} \\
    \text{s.t.}\ \theta^*(\lambda) &= \argmin_{\theta\in\R^d}g(\lambda, \theta) := \sum_{i=1}^{m}\lambda_i f_i(\theta)  &\text{Lower Level (LL)}. \label{opt: bo_ll}
\end{align}
Intuitively, the upper-level problem aims at searching the task weights $\lambda$ that optimizes the worst-case decrement, provided that the model parameters are the optimal point of the linearly-combined loss in the lower-level problem.

Note that the computation of $\nabla_{\lambda}\Phi(\lambda, \rho)$ involves the optimal solution of the lower-level optimization problem defined in \eqref{opt: bo_ll}, which typically requires estimating the Hessian inverse of the lower-level objective $g(\lambda, \theta)$. This can be done via Neumann series method~\citep{ghadimi2018approximation} or \texttt{SGD} based method~\citep{dagreou2022framework} when $g(\lambda, \theta)$ is strongly-convex in $\theta$. However, $f_i(\theta)$ is highly non-convex in MTL, making both \eqref{opt: bo_ul} and \eqref{opt: bo_ll} non-convex and cannot be solved in the worst-case scenario. We hence consider solving a surrogate problem. Suppose we have obtained $\lambda^k, \rho^k, \theta^k$ at iteration $k$. For \eqref{opt: bo_ul} we replace $\theta^*(\lambda)$ with $\theta^k$, which results in the following problem.
\begin{align}\label{opt: minmax_surro}
    \min_{\lambda\in \Delta^m}\max_{\rho\in \Delta^m}\ \Phi_k(\lambda, \rho) := \sum_{i=1}^{m}\rho_i(f_i(\cA(\lambda, \theta^k)) - f_i(\theta^k)),
\end{align}
whose gradients are given by
\begin{align}
    \nabla_{\lambda}\Phi_k(\lambda, \rho) &= \sum_{i=1}^{m}\rho_i\nabla_{\lambda}f_i(\cA(\lambda, \theta^k)), \label{eq: hypergrad_lam}\\
    \nabla_{\rho}\Phi_k(\lambda, \rho) &= \left(f_1(\cA(\lambda, \theta^k)) - f_1(\theta^k), ..., f_m(\cA(\lambda, \theta^k)) - f_m(\theta^k)\right)^\top. \label{eq: hypergrad_rho}
\end{align}
We can thus adopt the state-of-the-art bi-level optimization algorithms~\citep{chen2021closing, dagreou2022framework, chen2023optimal, hao2023bilevel} to perform the updates for $\lambda^k, \rho^k, \theta^k$ alternatingly: 
\begin{align*}
    (\lambda^{k+1},\ \rho^{k+1}) = \texttt{Optim}(\lambda^k, \rho^k, g^k),\
    \theta^{k+1} = \cA(\lambda^{k+1}, \theta^k),
\end{align*}
where $g^k$ denotes the stochastic gradients (with respect to $\lambda$ and $\rho$) of $\Phi_k$ in \eqref{opt: minmax_surro}. $\texttt{Optim}$ denotes the optimizer that performs one-step update for $(\lambda^k, \rho^k)$ to solve the min-max problem in~\eqref{opt: minmax_surro}, and $\cA(\lambda^{k+1}, \theta^k)$ denotes the optimizer for model training. For example, \texttt{Optim} and $\cA$ can be stochastic gradient descent ascent (\texttt{SGDA})~\citep{lin2020gradient} and \texttt{Adam}~\citep{kingma2014adam} respectively.

\subsection{Hypergradient estimation via zeroth-order method}\label{sec:hypergrad_zo_grad}
Before we present the algorithms that combine the strategies in Section \ref{sec: mtl_criterion} and \ref{sec: mtl_bo}, we further provide more details on the estimation of hypergradient $\nabla_{\lambda}\Phi_k(\lambda, \rho)$. 

We first observe that computing the hypergradient does not require computing $m$ gradients. Instead, with the help of auto differentiation, one may first construct the computational graph for the weighted loss $\sum_{i=1}^{m}\rho_i(f_i(\cA(\lambda, \theta)) - f_i(\theta))$, and then computes its gradient with respect to $\lambda$. We note that this computation involves computing gradients through the optimizer $\cA(\lambda, \theta)$, which can be efficiently calculated, for example, in PyTorch~\citep{paszke2019pytorch} or JAX~\citep{jax2018github} framework, by using strategies such as implicit differentiation and zeroth-order differentiation that are well supported in numerous differentiable optimization libraries~\citep{grefenstette2019generalized, deepmind2020jax, deleu2019torchmeta, arnold2020learn2learn, blondel2022efficient, grazzi2020iteration, choe2022betty, ren2023torchopt}. To further reduce the computational cost, we adopt the zeroth-order method based on the following lemma, whose proof is standard in zeroth-order optimization (see, e.g., Lemma A.1 in \cite{chen2022improve}). 
\begin{lemma}\label{lem: ZO_error}
    Suppose $f(\theta):\R^d \rightarrow \R$ is differentiable and $\ell$-smooth, i.e., $\norm{\nabla f(\theta_1) - \nabla f(\theta_2)}\leq \ell \norm{\theta_1 - \theta_2}$ for any $\theta_1, \theta_2\in\R^d$. Define the closed unit ball $\B_d:=\{u\in \R^d: \norm{u}\leq 1\}$ and the unit sphere $\bS_{d-1}:=\{u\in\R^d:\norm{u}=1\}$. Denote by $\textbf{Unif}$ the uniform distribution, and for $r>0$ and $v\in\R^d$ we define $f_r(\theta) = \E_{u\sim \text{Unif}(\B_d)}\left[f(\theta + ru)\right],\  G_{f}(\theta;r, v) := \frac{d}{r}f(\theta + rv)v$.
    Then we have
    \begin{align}
        \E_{v\sim \text{Unif}(\bS_{d-1})}\left[G_{f}(\theta;r, v)\right] = \nabla f_r(\theta),\  |f_r(\theta) - f(\theta)|\leq \frac{\ell r^2}{2},\ \norm{\nabla f_r(\theta) - \nabla f(\theta)}\leq \ell r. \label{eq: zo_bound_spt}
    \end{align}
\end{lemma}

\textbf{Remark.}\quad Note that we define the perturbed function value $f_r(\theta)$ and its associated zeroth-order gradient $G_{f}(\theta;r, v)$. Moreover, we can see from \eqref{eq: zo_bound_spt} that for $u\sim \text{Unif}(\B_d),\ v\sim \text{Unif}(\bS_{d-1})$, $f(\theta + ru)$ and $G_{f}(\theta;r, v)$ are biased estimate of $f(\theta)$ and $\nabla f(\theta)$ respectively, in which the biases can be well controlled by the perturbation scale $r$ and the smoothness constant $\ell$. We will thus utilize zeroth-order method to approximate the hypergradient $\nabla_{\lambda}\Phi_k(\lambda, \rho)$, whose details are summarized in Algorithm \ref{alg: hyper_marigold}. We note that zeroth-order optimization techniques have been studied extensively in optimization literature~\citep{wang2018stochastic, balasubramanian2018zeroth, wang2020zeroth, cai2022zeroth, balasubramanian2022zeroth, aghasi2024fully}, and were recently introduced to other domains of machine learning~\citep{malladi2024fine, zhang2024revisiting, zhang2023dpzero, tang2024private, yu2024privacy} for memory-efficient training.

\subsection{Algorithms}\label{sec: algorithm}
We are now ready to present our algorithms. In Algorithm \ref{alg: marigold}, we first estimate the hypergradient via Algorithm \ref{alg: hyper_marigold}, in which we adopt the single-point zeroth-order optimization method for efficient computation. We directly use the perturbed function value $f_i(\cA(\lambda + ru, \theta); \cB)$ in estimating $\nabla_{\rho}\Phi_k(\lambda, \rho)$ in \eqref{eq: hypergrad_rho} to save compute. On top of $g^k$, we perform the model tuning step for task weights $\lambda^k$ and $\rho^k$ via optimizer $\texttt{Optim\_UL}$. Finally we update model parameters $\theta^k$ by using any user-specific optimizer $\texttt{Optim\_LL}$.

\begin{algorithm}[ht]
\caption{\texttt{Hypergrad}}
\label{alg: hyper_marigold}
\begin{algorithmic}[1]
    \State \textbf{Input:} Model parameters $\theta$, task weights $\lambda, \rho$, perturbation scale $r$.
    \State Sample a batch of data $\cB$ and $u\sim \text{Unif}(\bS_{d-1})$. Compute
    \begin{align*}
        \widetilde{\nabla_{\lambda} \Phi_k}(\lambda, \rho;\cB) &=\frac{m}{r}\left(\sum_{i=1}^{m}\rho_if_i(\cA(\lambda + ru, \theta);\cB)\right)u, \\
        \widetilde{\nabla_{\rho} \Phi_k}(\lambda, \rho;\cB) &=  \left(f_1(\cA(\lambda + ru, \theta);\cB) - f_1(\theta;\cB), ..., f_m(\cA(\lambda + ru, \theta);\cB) - f_m(\theta;\cB)\right)^\top.
    \end{align*}
    \State \textbf{Output:} $g^k = (\widetilde{\nabla_{\lambda} \Phi_k}(\lambda, \rho;\cB), \widetilde{\nabla_{\rho} \Phi_k}(\lambda, \rho;\cB))$.
\end{algorithmic}
\end{algorithm}


\begin{algorithm}[ht]
\caption{\texttt{MARIGOLD}: Multi-task gradient balancing via zeroth-order bi-level differentiation}
\label{alg: marigold}
\begin{algorithmic}[1]
    \State \textbf{Input:} Initial model parameters $\theta^0$, task weights $\lambda^0, \rho^0$, perturbation scale $r$.
    \For {$k=0, \dots, K-1$}
    \State Estimate the hypergradient
        \[
            g^{k} = \texttt{Hypergrad}(\theta^k, \lambda^k, \rho^k, r).
        \]
    \State Task weights tuning 
        \[
            \lambda^{k+1}, \rho^{k+1} = \texttt{Optim\_UL}(\lambda^k, \rho^k, g^k).
        \]
    \State Model training
        \[
            \theta^{k+1} = \texttt{Optim\_LL}(\lambda^{k+1}, \theta^{k}).
        \]
    \EndFor
    \State \textbf{Output:} $\theta^{K}$.
\end{algorithmic}
\end{algorithm}
Our algorithm is model-agnostic, in the sense that the algorithm $\cA(\lambda, \theta)$ (i.e., $\texttt{Optim\_LL}$ in Algorithm \ref{alg: marigold}) for model training is not restricted to \texttt{SGD}, and does not affect the choice of optimizers of the task weights (i.e., $\texttt{Optim\_UL}$). 

We note that the practical implementation of MTL experiments usually chooses $\cA(\lambda, \theta)$ to be the \texttt{Adam} optimizer~\citep{liu2021conflict, navon2022multi, xiao2023direction, liu2024famo}, leading to an inconsistency with what their theory suggests. We point out that the reason of this mismatch can be rooted back to the linearization step in \eqref{eq: linearization}, which limits the set of available optimizer candidates for $\cA$. 
\subsection{Beyond worst-case decrement -- a general framework}\label{sec: general_framework}
Now that we have seen from previous Sections the gradient balancing techniques like \texttt{CAGrad} can be reformulated as a bi-level optimization problem, we can further generalize our framework to handle other problems with different objectives. In this section we aim at tuning task weights based on a user-specified function $R(\lambda, \theta)$, which leads to the following problem.
\begin{align*}
    \min_{\lambda\in \Delta^m}\ R(\lambda, \theta^*(\lambda)),\quad \text{s.t.}\ \theta^*(\lambda) = \argmin_{\theta\in\R^d}g(\lambda, \theta). 
\end{align*}
Similar to \eqref{opt: minmax_surro}, we may consider the surrogate of the above problem as follows.
\begin{align*}
    \min_{\lambda\in\Delta^m}R(\lambda, \theta^k)
\end{align*}
where $\theta^{k+1} = \cA(\lambda^k, \theta^k)$. Setting $R(\lambda, \theta)$ as the worst-case decrement in \eqref{eq: worst_case_decrement} recovers Algorithm \ref{alg: marigold} for MTL problems. We may also define $R(\lambda, \theta)$ as the loss function of certain main tasks, and we can recover auxiliary learning problems~\citep{jaderberg2016reinforcement, goyal2019scaling, liu2019self, navon2020auxiliary}. We will test the performance of our algorithmic framework on these two types of problems in Section \ref{sec: exp}.

\section{Experiments}\label{sec: exp}
To empirically demonstrate the performance of our Algorithm \ref{alg: marigold}, we conduct experiments on both public datasets widely used in MTL literature and industrial-scale datasets. The experimental details are deferred to Section \ref{sec: exp_app} in the Appendix.

\subsection{Public data}\label{sec: public_exp}
\textbf{Algorithms.}\quad We investigate the performance of different algorithms on public datasets. Following the multi-task learning problems in computer vision~\citep{liu2019end, liu2021conflict, liu2024famo, xiao2023direction}, we adopt SegNet, an encoder-decoder network for semantic segmentation and depth estimation developed in~\cite{liu2019end}, to test the performance of our Algorithms on the NYU-v2 dataset~\citep{silberman2012indoor} and the CityScapes dataset~\citep{cordts2016cityscapes}. In addition to the Single-Task Learning (\texttt{STL}), which trains each individual task separately, we also include existing state-of-the-art MTL algorithms with gradient balancing techniques for a thorough comparison:  Multiple-Gradient Descent Algorithm (\texttt{MGDA})~\citep{sener2018multi}, Projecting Conflicting Gradients (\texttt{PCGrad})~\citep{yu2020gradient}, Gradient Sign Dropout (\texttt{GradDrop})~\citep{chen2020just}, Conflict-Averse Gradient descent (\texttt{CAGrad})~\citep{liu2021conflict}, Impartial Multi-Task Learning-Grad (\texttt{IMTL-G})~\citep{liu2021towards}, Multi-objective gradient with Correction (\texttt{MoCo})~\citep{fernando2022mitigating}, Multi-objective gradient with Double sampling (\texttt{MoDo})~\citep{chen2023three}, Nash bargaining Multi-Task Learning (\texttt{Nash-MTL})~\citep{navon2022multi}, Fast Adaptive Multitask Optimization (\texttt{FAMO})~\citep{liu2024famo}, Stochastic Direction-oriented Multi-objective Gradient descent (\texttt{SDMGrad})~\citep{xiao2023direction}. We implement Algorithm \ref{alg: marigold} (\texttt{MARIGOLD}) based one the codebase released by \cite{liu2021conflict} and contributed by, e.g., \cite{navon2022multi}, \cite{liu2024famo}, and \cite{xiao2023direction}.

In addition to the gradient balancing methods, we include the following loss balancing methods in Section \ref{sec: exp_app}: Linear Scalarization (\texttt{LS}), which optimizes the sum of all task losses; Scale-Invariant (\texttt{SI}), which minimizes the sum of the log of task losses; Random Loss Weighting (\texttt{RLW})~\citep{lin2021closer}, Dynamic Weight Average (\texttt{DWA})~\citep{liu2019end}, Uncertainty Weighting (\texttt{UW})~\citep{kendall2018multi}.

\textbf{Metrics.}\quad There are in total three tasks including image segmentation, depth prediction, and surface normal prediction. On Cityscapes dataset we aim at solving the first two tasks and for NYU-v2 we solve all three tasks. In the table we adopt the metrics used in previous works~\citep{liu2019end, liu2021conflict, liu2024famo, xiao2023direction}, and present them in Tables \ref{tab:nyuv2_grad} and \ref{tab:cityscapes_grad}. Note that for the last three columns of the tables, Mean Rank (MR) refers to the average of the rank of the algorithm on all tasks, $\Delta k\% = \frac{1}{m}\sum_{i=1}^m (-1)^{\delta_i} (M_{k,i} - M_{b,i}) / M_{b,i} \times 100 $ where $M_{b,i}$ and $M_{k,i}$ are the STL and algorithm $k$'s value for metric $M_i$. $\delta_i = 1$ (or $0$) if the $M_i$ is higher (or lower) the better, and ``Cost'' represents the per-iteration computational cost of the corresponding algorithm. These three metrics all evaluate the performance and efficiency of MTL algorithms. All metrics are computed by averaging the results in the last ten epochs and further averaged over three random seeds. 

\begin{figure}[t]
    \centering
    \subfigure[Cityscapes]{\includegraphics[width=0.45\textwidth]{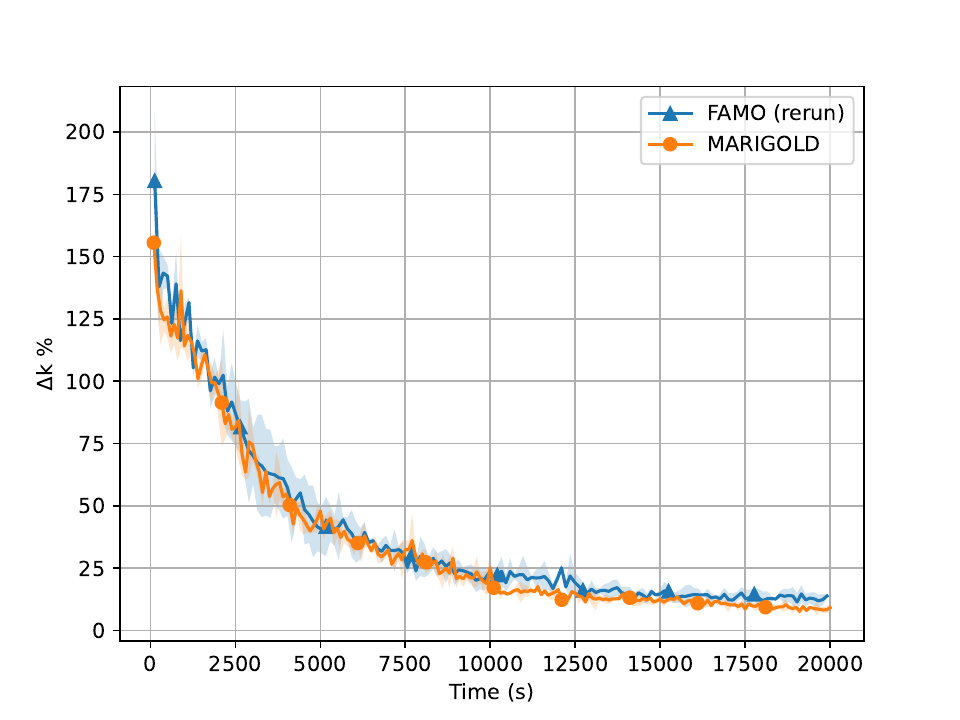}\label{fig: city_delta}}
    \subfigure[NYU-v2]{\includegraphics[width=0.45\textwidth]{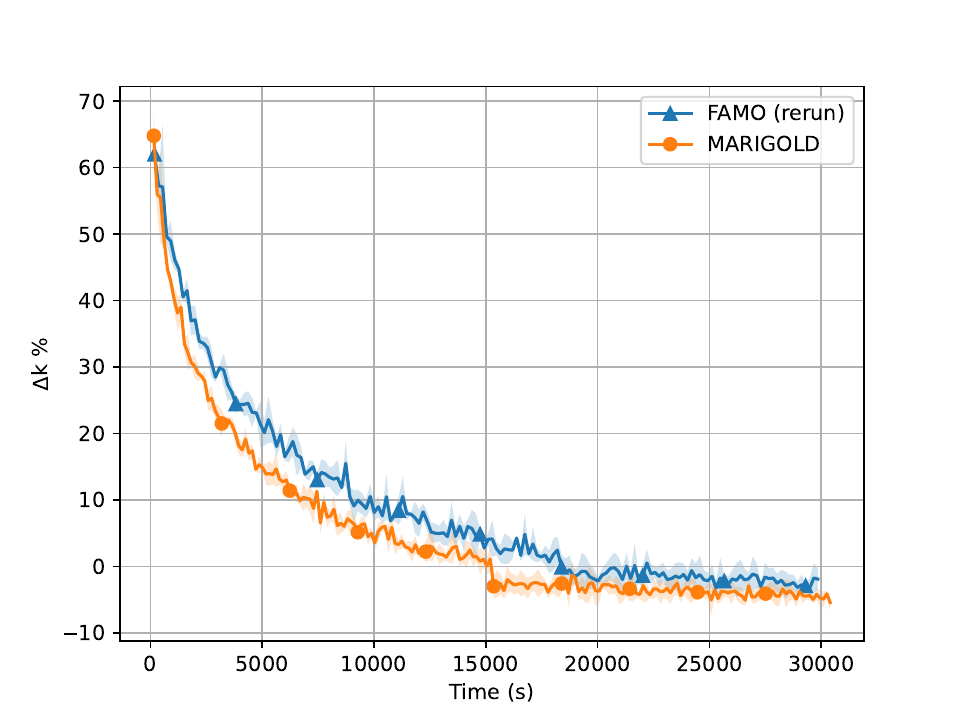}\label{fig: nyu_delta}}
    \caption{Comparison between \texttt{FAMO} and Algorithm \ref{alg: marigold} in terms of $\Delta k\%$ over time.}
    \label{fig: delta_curves}
\end{figure}

In Table \ref{tab:time}, we further compare the wall-clock time per training epoch of our Algorithm \ref{alg: marigold} with the most efficient gradient balancing method \texttt{FAMO}, which is reported~\citep{liu2024famo} to be much more efficient than \texttt{MGDA}, \texttt{PCGrad}, \texttt{CAGrad}, \texttt{IMTL-G}, and \texttt{NashMTL}. We also include the time of \texttt{MGDA} as a representative of these less-efficient ones. We do not include \texttt{SDMGrad}~\citep{xiao2023direction} as it shares similarity with \texttt{MGDA}-type methods, and in the official implementation~\citep{xiao2023direction} one epoch of training typically takes around 30 minutes, which is much longer than \texttt{MGDA}. Moreover, we showcase how $\Delta k \%$ of \texttt{FAMO} and \texttt{MARIGOLD}, two most efficient and competitive algorithms in Tables \ref{tab:nyuv2_grad} and \ref{tab:cityscapes_grad}, evolve over time in Figure \ref{fig: delta_curves}, in which the shaded region represents the standard deviation. Note that to plot Figure \ref{fig: delta_curves}, we rerun \cite{liu2024famo} on our device for three trials, and include results in Tables \ref{tab:nyuv2_grad}, \ref{tab:cityscapes_grad}, \ref{tab:time} and Figure \ref{fig: delta_curves} named ``\texttt{FAMO} (rerun)".

\textbf{Observations.}\quad We can observe from Tables 
\ref{tab:nyuv2_grad}, \ref{tab:cityscapes_grad}, and \ref{tab:time} and Figure \ref{fig: delta_curves} that gradient balancing methods can be divided into two groups, one with $\cO(md)$ per-iteration cost and the other one with $\cO(d)$ cost only, and our Algorithm \texttt{MARIGOLD} is among the most efficient ones. In particular, we notice that: (i) \textbf{Under same number of epochs}, we note from Tables \ref{tab:nyuv2_grad} and \ref{tab:cityscapes_grad} that the performance of our algorithm is better than or comparable with existing ones; (ii) \textbf{Under same amount of time}, the results in Figure \ref{fig: delta_curves} validate the overall performance gains of \texttt{MARIGOLD} over \texttt{FAMO}.

\begin{table}[t]
\caption{Comparison between gradient balancing methods and Algorithm \ref{alg: marigold} on NYU-v2 dataset.}
\label{tab:nyuv2_grad}
  \centering
  \begin{adjustbox}{max width=\textwidth}
  \begin{tabular}{lcccccccccccc}
    \toprule
    \multirow{3}*{Method} & \multicolumn{2}{c}{Segmentation} & \multicolumn{2}{c}{Depth} & \multicolumn{5}{c}{Surface Normal} & \multirow{3}*{MR $\downarrow$} & \multirow{3}*{$\Delta k\%\downarrow$} & \multirow{3}*{Cost} \\
    \cmidrule(lr){2-3}\cmidrule(lr){4-5}\cmidrule(lr){6-10}
    & \multirow{2}*{mIoU $\uparrow$} & \multirow{2}*{Pix Acc $\uparrow$} & \multirow{2}*{Abs Err $\downarrow$} & \multirow{2}*{Rel Err $\downarrow$} & \multicolumn{2}{c}{Angle Distance $\downarrow$} & \multicolumn{3}{c}{Within $t^\circ$ $\uparrow$} & \\
    \cmidrule(lr){6-7}\cmidrule(lr){8-10}
    & & & & & Mean & Median & 11.25 & 22.5 & 30 & \\
    \midrule
    \texttt{STL} & 38.30 & 63.76 & 0.6754 & 0.2780 & 25.01 & 19.21 & 30.14 & 57.20 & 69.15 & \\
    \midrule
    \texttt{MGDA}~\citep{desideri2012multiple} & 30.47 & 59.90 & 0.6070 & 0.2555 & 24.88 & 19.45 & 29.18 & \textbf{56.88} & \textbf{69.36} & 6.22 & 1.38 & $\cO(md)$\\
    \texttt{PCGrad}~\citep{yu2020gradient} & 38.06 & 64.64 & 0.5550 & 0.2325 & 27.41 & 22.80 & 23.86 & 49.83 & 63.14 & 10.33 & 3.97 & $\cO(md)$\\
    \texttt{GradDrop}~\citep{chen2020just} & 39.39 & 65.12 & 0.5455 & 0.2279 & 27.48 & 22.96 & 23.38 & 49.44 & 62.87 & 9.78 & 3.58 & $\cO(md)$\\
    \texttt{CAGrad}~\citep{liu2021conflict} & 39.79 & 65.49 & 0.5486 & 0.2250 & 26.31 & 21.58 & 25.61 & 52.36 & 65.58 & 7.89 & 0.20 & $\cO(md)$\\
    \texttt{IMTL-G}~\citep{liu2021towards} & 39.35 & 65.60 & 0.5426 & 0.2256 & 26.02 & 21.19 & 26.20 & 53.13 & 66.24 & 7.11 & -0.76 & $\cO(md)$\\
    \texttt{MoCo}~\citep{fernando2022mitigating} & 40.30 & 66.07 & 0.5575 & 0.2135 & 26.67 & 21.83 & 25.61 & 51.78 & 64.85 & 7.44 & 0.16 & $\cO(md)$\\
    \texttt{MoDo}~\citep{chen2023three} & 35.28 & 62.62 & 0.5821 & 0.2405 & 25.65 & 20.33 & 28.04 & 54.86 & 67.37 & 8.44 & 0.49 & $\cO(md)$\\
    \texttt{Nash-MTL}~\citep{navon2022multi} & 40.13 & 65.93 & 0.5261 & 0.2171 & 25.26 & 20.08 & 28.40 & 55.47 & 68.15 & 4.33 & -4.04 & $\cO(md)$\\
    \texttt{SDMGrad}~\citep{xiao2023direction} & 40.47 & 65.90 & 0.5225 & \textbf{0.2084} & 25.07 & 19.99 & 28.54 & 55.74 & 68.53 & \textbf{3.00} & \textbf{-4.84} & $\cO(md)$\\
    \midrule
    \texttt{FAMO}~\citep{liu2024famo} & 38.88 & 64.90 & 0.5474 & 0.2194 & 25.06 & 19.57 & \textbf{29.21} & 56.61 & 68.98 & 4.44 & -4.10 & $\cO(d)$\\
    \texttt{FAMO} (rerun)~\citep{liu2024famo} &38.60 & 63.29 & \textbf{0.5201} & 0.2185 & \textbf{24.50} & \textbf{19.36} & 27.48 &  53.75& 65.91& 5.44 & -3.27 & $\cO(d)$ \\
    \texttt{MARIGOLD} & \textbf{41.01}& \textbf{66.27} & 0.5270 & 0.2137  & 25.29  & 20.06 & 28.55 & 55.51 & 68.11  & 3.56 & -4.54& $\cO(d)$\\
    \bottomrule
  \end{tabular}
  \vspace{-0.2cm}
  \end{adjustbox}
\end{table}

\begin{table}[t]
\caption{Comparison between gradient balancing methods and Algorithm \ref{alg: marigold} on Cityscapes dataset.}
\label{tab:cityscapes_grad}
  \centering
    \begin{adjustbox}{max width=0.7\textwidth}
  \begin{tabular}{lccccccc}
    \toprule
    \multirow{2}*{Method} & \multicolumn{2}{c}{Segmentation} & \multicolumn{2}{c}{Depth} & 
    \multirow{2}*{MR $\downarrow$} &
    \multirow{2}*{$\Delta k\%\downarrow$} & \multirow{2}*{Cost} \\
    \cmidrule(lr){2-3}\cmidrule(lr){4-5}
    & mIoU $\uparrow$ & Pix Acc $\uparrow$ & Abs Err $\downarrow$ & Rel Err $\downarrow$ & \\
    \midrule
    \texttt{STL} & 74.01 & 93.16 & 0.0125 & 27.77 & \\
    \midrule
    \texttt{MGDA}~\citep{desideri2012multiple} & 68.84 & 91.54 & 0.0309 & 33.50 & 9.50 & 44.14 & $\cO(md)$ \\
    \texttt{PCGrad}~\citep{yu2020gradient} & 75.13 & 93.48 & 0.0154 & 42.07 & 8.50 & 18.29 & $\cO(md)$ \\
    \texttt{GradDrop}~\citep{chen2020just} & 75.27 & 93.53 & 0.0157 & 47.54 & 7.50 & 23.73 & $\cO(md)$ \\
    \texttt{CAGrad}~\citep{liu2021conflict} & 75.16 & 93.48 & 0.0141 & 37.60 & 7.00 & 11.64 & $\cO(md)$ \\
    \texttt{IMTL-G}~\citep{liu2021towards} & 75.33 & 93.49 & 0.0135 & 38.41 & 5.00 & 11.10 & $\cO(md)$\\
    \texttt{MoCo}~\citep{fernando2022mitigating} & \textbf{75.42} & 93.55 & 0.0149 & 34.19 & 3.75 & 9.90 & $\cO(md)$\\
    \texttt{MoDo}~\citep{chen2023three} & 74.55 & 93.32 & 0.0159 & 41.51 & 9.75 & 18.89 & $\cO(md)$\\
    \texttt{Nash-MTL}~\citep{navon2022multi} & 75.41 & \textbf{93.66} & \textbf{0.0129} & 35.02 & \textbf{2.50} & \textbf{6.82}& $\cO(md)$ \\
    \texttt{SDMGrad}~\citep{xiao2023direction} & 74.53 & 93.52 & 0.0137 & 34.01 & 5.25 & 7.79 & $\cO(md)$\\
    \midrule
    \texttt{FAMO}~\citep{liu2024famo} & 74.54 & 93.29 & 0.0145 & \textbf{32.59} & 7.00 & 8.13 & $\cO(d)$\\
    \texttt{FAMO} (rerun)~\citep{liu2024famo} & 74.76 & 93.25 & 0.0140 & 35.78 & 7.75 & 9.92 & $\cO(d)$ \\
    \texttt{MARIGOLD} & 75.30 & 93.51 & 0.0139 & 34.79 & 4.50 & 8.62 & $\cO(d)$\\
    \bottomrule
  \end{tabular}
  \end{adjustbox}
\end{table}

\begin{table}[t]
\caption{Wall-clock time (s) per epoch comparison between \texttt{FAMO} and Algorithm \ref{alg: marigold}.}
\label{tab:time}
\centering
\begin{tabular}{ccc}
\toprule
Method & NYU-v2 & Cityscapes  \\ 
  \midrule
  \texttt{MGDA}   & 375 & 163  \\
\texttt{FAMO} (rerun)    & 182  & 126  \\ 
\texttt{MARIGOLD} & 152 & 100  \\ 
\bottomrule
\end{tabular}
\vspace{-0.3cm}
\end{table}

\subsection{Industrial-scale data}\label{sec: industry_exp}
To further demonstrate the superior performance of our methods on industrial-scale models and datasets, we conduct experiments on a Meta's large foundation model, which is built on top of the advanced architectures, such as \cite{zhang2022dhen} and \cite{zhang2024wukong}, for solving large-scale ads ranking problems. In particular, we consider the auxiliary learning problem, a special case of the framework in Section \ref{sec: general_framework}.
\begin{align}
    &\min_{\omega \in \R}\ f_{\text{click}}(\theta^*(\omega)), \label{opt: industry_ul} \\    
    &\text{s.t.}\ \theta^*(\omega) = \argmin_{\theta\in \R^d} f(\omega, \theta) := f_{\text{click}}(\theta) + f_{\text{conv|click}}(\theta) + f_{\text{conv}}(\theta) + \omega f_{\text{distill}}(\theta), \label{opt: industry_ll}
\end{align}
where we have four different tasks. The main tasks include post-view click (CTR), post-click conversion (Conv|click) and conversion (CVR), which are standard in ads ranking problems~\citep{ma2018entire, wen2020entire}. Their loss functions are represented by $f_{\text{click}}, f_{\text{conv|click}}, f_{\text{conv}}$ respectively. We further introduce one auxiliary task $f_{\text{distill}}$, which is acquired by leveraging the knowledge distilled from a larger teacher model, to enhance the training of main tasks. We set the upper-level objective function to be $f_{\text{click}}$ as we aim at improving the model performance on predicting the post-view click rate. Similar to \ref{opt: bo_ll}, $\theta^*(\omega)$ is often intractable as $f(\omega, \theta)$ is in general non-convex. Based the idea in Section \ref{sec: mtl_bo}, we consider the following problem as a surrogate.
\begin{align*}
    \min_{\omega} f_{\text{click}}(\cA(\omega, \theta^k))
\end{align*}
where $\cA(\omega, \theta^k)$ denotes the training algorithm for optimizing $f(\omega, \theta)$. To update $\omega^k$ at $k$-th iteration, we will compute $\nabla_{\omega}f_{\text{click}}(\cA(w^k, \theta^k))$ and run algorithms like gradient descent or adaptive methods. We update the model parameters $\theta^k$ using the default optimizer \texttt{Adam}.

\textbf{Metrics.}\quad We use the Normalized Entropy (NE) metric~\citep{he2014practical}, which is common in ads ranking problems. The lower NE is, the better. We evaluate NE for both \texttt{LS} and \texttt{MARIGOLD}, and the NE gains of \texttt{MARIGOLD} over \texttt{LS} on all tasks are summarized in Table \ref{tab:industry}.

\begin{table*}[ht]
\caption{NE gains of our algorithm \ref{alg: marigold} over \texttt{LS} (see Section \ref{sec: public_exp}) on all tasks.}
    \label{tab:industry}
\begin{center}
    \begin{tabular}{ccccc}
    \toprule
    Task    & Click  & Conv|click & Conv   & Distillation \\ 
    \midrule
    NE gain & 0.08\% & 0.03\%     & 0.07\% & 0.14\%       \\ 
    \bottomrule
    \end{tabular}
\end{center}
\end{table*}

\textbf{Observations.}\quad From Table \ref{tab:industry} we can find that clearly our Algorithm \ref{alg: marigold} is much better than the baseline setting which assigns equal weights to all tasks and does not conduct any tuning. This further demonstrates the potential of our methods in industrial-scale multi-task learning problems.

\section{Conclusion}\label{sec: conclusion}
We propose a novel algorithm for efficiently solving multi-task gradient balancing problems in MTL. Different from most existing algorithms that require expensive compute and memory to handle task gradients, our method only requires zeroth-order information to estimate the hypergradient of the bi-level optimization problem, which further captures the critical interactions needed for balancing. Moreover, we empirically verify the effectiveness of our approach on both public dataset and industrial dataset. There are several interesting questions unanswered. For example, what is the convergence rate of our algorithm for finding a Pareto stationary point, would the techniques in our paper be useful to accelerate the training of other types of bi-level optimization problems like meta learning and reinforcement learning. We leave them as the future work.

\newpage


\appendix

\newpage

\appendix

\section{Related Work}\label{sec: related_work_app}

{\bf Multi-task learning and multi-objective optimization.}\quad In machine learning community, multi-task learning (MTL)~\citep{caruana1997multitask} often refers to the training technique that combines and solves different tasks in a unified model. Specifically, the model usually contains two parts -- parameters that are shared across different tasks (shared parameters), and parameters that are designed for specific tasks (task-specific parameters). Multi-objective optimization (MOO) originates from optimization community~\citep{miettinen1999nonlinear}. Different from single-objective optimization, one major challenge in MOO is to efficiently find the Pareto optimal/stationary points of multiple objective functions through multiple gradient descent algorithm (\texttt{MGDA}). In recent years, the boundary between MTL and MOO becomes blurred and a huge amount of efforts have been dedicated to understand MTL through the lens of MOO~\citep{sener2018multi, liu2021conflict, liu2021stochastic, zhou2022convergence, fernando2022mitigating, chen2023three, xiao2023direction}, which provides solid theoretical convergence guarantees.

In addition to MOO, another line of work propose different strategies to perform other types of gradient manipulation to achieve better performance, such as \texttt{GradNorm}~\citep{chen2018gradnorm}, \texttt{PCGrad}~\citep{yu2020gradient}, \texttt{GradDrop}~\citep{chen2020just}, \texttt{GradVac}~\citep{wang2021gradient}, etc. It is worth noting that, gradient balancing that requires the knowledge of different task gradients is notoriously time-consuming, as it often requires computing and storing $m$ different task gradients. Motivated by this, some works apply several approximation strategies are dedicated to mitigate this issue. For example, \cite{sener2018multi} considers gradient balancing applied to representation level gradients, \cite{liu2024famo} uses loss balancing to approximate the gradient balancing. However, it remains unclear if the original \texttt{MGDA}-type methods can be efficiently implemented without any approximation.

Another closely related problem is auxiliary learning, in which we have two different classes of tasks -- auxiliary tasks and main tasks. We do not care about the performance of auxiliary tasks as they are introduced to only benefit the training of main ones~\citep{jaderberg2016reinforcement, goyal2019scaling, liu2019self, navon2020auxiliary}.

{\bf Bi-level optimization.}\quad The study of bi-level optimization can be dated back to~\cite{stackelberg1952theory, bracken1973mathematical}. Recently, bi-level optimization is gaining popularity in both machine learning and optimization community, due to its capabilities of handling problems with a nested structure, such as hyperparameter optimization~\citep{lorraine2020optimizing}, reinforcement learning~\citep{hong2023two}, meta learning~\citep{rajeswaran2019meta}, model pruning~\citep{zhang2022advancing}, min-max optimization~\citep{chen2021closing}, etc. Another line of work is dedicated to settle the convergence rate as well as the sample complexity of bi-level optimization algorithms from a theoretical perspective~\citep{ghadimi2018approximation, hong2023two, chen2021closing, dagreou2022framework, chen2023optimal, hao2023bilevel}. It has been observed that MTL is closely related to meta learning and thus the intersection of MTL and bi-level optimization has been explored in different aspects~\citep{navon2020auxiliary, wang2021bridging}. As we shall see in Section~\ref{sec: method}, many MTL algorithms have an intrinsic bi-level structure, which can be efficiently solved via state-of-the-art ideas in bi-level optimization.

\section{Complexity of gradient balancing algorithms in Tables \ref{tab:nyuv2_grad} and \ref{tab:cityscapes_grad}}\label{sec: grad_balance}
In this section we give a brief introduction of gradient balancing methods in Tables \ref{tab:nyuv2_grad} and \ref{tab:cityscapes_grad} and discuss their per-iteration computational cost. Recall that we use $m$ to denote the number of tasks and $d$ to represent the dimension of model parameters $\theta$.

\textbf{MGDA-type methods.}\quad The Multiple-Gradient Descent Algorithm (\texttt{MGDA})~\citep{sener2018multi} and its variants, including Conflict-Averse Gradient descent (\texttt{CAGrad})~\citep{liu2021conflict}\footnote{It has been shown in \cite{xiao2023direction} that \texttt{CAGrad} and \texttt{SDMGrad} are closely related to \texttt{MGDA}, and thus we include them in this category for simplicity.}, Multi-objective gradient with Correction (\texttt{MoCo})~\citep{fernando2022mitigating}, Multi-objective gradient with Double sampling (\texttt{MoDo})~\citep{chen2023three} and Stochastic Direction-oriented Multi-objective Gradient descent (\texttt{SDMGrad})~\citep{xiao2023direction}, first compute all task gradients, and then find the convex combination of all gradients with the minimum norm. These methods require computing and storing all task gradients, which has $\cO(md)$ per-iteration computational cost.

\textbf{PCGrad.}\quad Projecting Conflicting Gradients (\texttt{PCGrad})~\citep{yu2020gradient}, also known as Gradient Surgery, performs some projection operations whenever a gradient conflict happens. To check if there is a conflict between pairwise gradients, it requires access to all task gradients, which has $\cO(md)$ per-iteration computational cost.

\textbf{GradDrop.}\quad Gradient Sign Dropout (\texttt{GradDrop})~\citep{chen2020just} constructs a sign purity measure at every gradient location for each task, leading to $\cO(md)$ per-iteration computational cost.

\textbf{IMTL-G.}\quad Impartial Multi-Task Learning-Grad (\texttt{IMTL-G})~\citep{liu2021towards} calculates the task weights by using task gradient differences and unit-norm gradient differences using all task gradients, which thus has $\cO(md)$ per-iteration computational cost.

\textbf{Nash-MTL.}\quad Nash bargaining Multi-Task Learning (\texttt{Nash-MTL})~\citep{navon2022multi} tackles the MTL problem from a game theory perspective, and solves the task weights $\lambda$ via a linear system $A\lambda = 1/\lambda$, where $A = (\<\nabla f_i(\theta;\cB),\nabla f_j(\theta;\cB)>)$ encodes the pairwise gradient similarity, whose computation has $\cO(md)$ per-iteration computational cost.

\textbf{FAMO.}\quad Fast Adaptive Multitask Optimization (\texttt{FAMO})~\citep{liu2024famo} estimates a matrix-vector product used in most gradient balancing methods via differences between losses (see Equation (9) in \citep{liu2024famo}), and thus can be thought of as an approximate gradient balancing method, which only has $\cO(d)$ (one backward pass of the weighted loss) per-iteration computational cost. However, the official implementation~\citep{liu2024famo} of \texttt{FAMO} applies \texttt{Adam} optimizer, while the theory suggests \texttt{SGD}, making it less explainable than our framework, which works for any optimizer.

\textbf{MARIGOLD.}\quad Multi-tAsk gRadIent balancinG via zerOth-order bi-leveL Differentiation (Algorithm \ref{alg: marigold}, \texttt{MARIGOLD}) treats the gradient balancing and the model training as a bi-level optimization problem. The only additional cost, as compared to vanilla model training, is the estimation of the hypergradient, which can be computed efficiently via single-point zeroth-order method in Algorithm \ref{alg: hyper_marigold}. Thus it only has $\cO(d)$ (one backward pass of the weighted loss) per-iteration computational cost.

\section{Experimental details}\label{sec: exp_app}

\begin{table}[!ht]
  \caption{Comparison between loss balancing methods and Algorithm \ref{alg: marigold} on NYU-v2 dataset.}
  \label{tab:nyuv2_loss}
  \centering
  \begin{adjustbox}{max width=\textwidth}
  \begin{tabular}{lccccccccccc}
    \toprule
    \multirow{3}*{Method} & \multicolumn{2}{c}{Segmentation} & \multicolumn{2}{c}{Depth} & \multicolumn{5}{c}{Surface Normal} & \multirow{3}*{MR $\downarrow$} & \multirow{3}*{$\Delta k\%\downarrow$} \\
    \cmidrule(lr){2-3}\cmidrule(lr){4-5}\cmidrule(lr){6-10}
    & \multirow{2}*{mIoU $\uparrow$} & \multirow{2}*{Pix Acc $\uparrow$} & \multirow{2}*{Abs Err $\downarrow$} & \multirow{2}*{Rel Err $\downarrow$} & \multicolumn{2}{c}{Angle Distance $\downarrow$} & \multicolumn{3}{c}{Within $t^\circ$ $\uparrow$} & \\
    \cmidrule(lr){6-7}\cmidrule(lr){8-10}
    & & & & & Mean & Median & 11.25 & 22.5 & 30 & \\
    \midrule
    \texttt{STL} & 38.30 & 63.76 & 0.6754 & 0.2780 & 25.01 & 19.21 & 30.14 & 57.20 & 69.15 & \\
    \midrule
    \texttt{LS} & 39.29 & 65.33 & 0.5493 & 0.2263 & 28.15 & 23.96 & 22.09 & 47.50 & 61.08 & 4.22 & 5.59  \\
    \texttt{SI} & 38.45 & 64.27 & 0.5354 & 0.2201 & 27.60 & 23.37 & 22.53 & 48.57 & 62.32 & 3.44 & 4.39 \\
    \texttt{RLW}~\citep{lin2021reasonable} & 37.17 & 63.77 & 0.5759 & 0.2410 & 28.27 & 24.18 & 22.26 & 47.05 & 60.62 & 5.67 & 7.78 \\
    \texttt{DWA}~\citep{liu2019end} & 39.11 & 65.31 & 0.5510 & 0.2285 & 27.61 & 23.18 & 24.17 & 50.18 & 62.39 & 3.33 & 3.57 \\
    \texttt{UW}~\citep{kendall2018multi} & 36.87 & 63.17 & 0.5446 & 0.2260 & 27.04 & 22.61 & 23.54 & 49.05 & 63.65 & 3.33 & 4.05 \\
    \midrule
    \texttt{MARIGOLD} & \textbf{41.01}& \textbf{66.27} & \textbf{0.5270} & \textbf{0.2137}  & \textbf{25.29}  & \textbf{20.06} & \textbf{28.55} & \textbf{55.51} & \textbf{68.11}  & \textbf{1.00} & \textbf{-4.54}\\
    \bottomrule
  \end{tabular}
  \end{adjustbox}
\end{table}

\begin{table}[!ht]
\caption{Comparison between loss balancing methods and Algorithm \ref{alg: marigold} on Cityscapes dataset.}
\label{tab:cityscapes_loss}
  \centering
    \begin{adjustbox}{max width=0.7\textwidth}
  \begin{tabular}{lcccccc}
    \toprule
    \multirow{2}*{Method} & \multicolumn{2}{c}{Segmentation} & \multicolumn{2}{c}{Depth} & 
    \multirow{2}*{MR $\downarrow$} &
    \multirow{2}*{$\Delta k\%\downarrow$} \\
    \cmidrule(lr){2-3}\cmidrule(lr){4-5}
    & mIoU $\uparrow$ & Pix Acc $\uparrow$ & Abs Err $\downarrow$ & Rel Err $\downarrow$ & \\
    \midrule
    \texttt{STL} & 74.01 & 93.16 & 0.0125 & 27.77 & \\
    \midrule
    \texttt{LS} & 75.18 & 93.49 & 0.0155 & 46.77 & 3.50 & 22.60  \\
    \texttt{SI} & 70.95 & 91.73 & 0.0161 & 33.83 & 5.00 & 14.11 \\
    \texttt{RLW}~\citep{lin2021reasonable} & 74.57 & 93.41 & 0.0158 & 47.79 & 4.50 & 24.38 \\
    \texttt{DWA}~\citep{liu2019end} & 75.24 & \textbf{93.52} & 0.0160 & 44.37 & 3.00 & 21.45 \\
    \texttt{UW}~\citep{kendall2018multi} & 72.02 & 92.85 & 0.0140 & \textbf{30.13} & 3.25 & \textbf{5.89} \\
    \midrule
    \texttt{MARIGOLD} & \textbf{75.30} & 93.51 & \textbf{0.0139} & 34.79 & \textbf{1.75} & 8.62 \\
    \bottomrule
  \end{tabular}
  \end{adjustbox}
\end{table}

\begin{table}[!ht]
\caption{Multi-task supervised learning on NYU-v2 dataset.}
\label{tab:nyuv2_app}
  \centering
  \begin{adjustbox}{max width=\textwidth}
  \begin{tabular}{llllllllllll}
    \toprule
    \multirow{3}*{Method} & \multicolumn{2}{c}{Segmentation} & \multicolumn{2}{c}{Depth} & \multicolumn{5}{c}{Surface Normal} & \multirow{3}*{MR $\downarrow$} & \multirow{3}*{$\Delta k\%\downarrow$} \\
    \cmidrule(lr){2-3}\cmidrule(lr){4-5}\cmidrule(lr){6-10}
    & \multirow{2}*{mIoU $\uparrow$} & \multirow{2}*{Pix Acc $\uparrow$} & \multirow{2}*{Abs Err $\downarrow$} & \multirow{2}*{Rel Err $\downarrow$} & \multicolumn{2}{c}{Angle Distance $\downarrow$} & \multicolumn{3}{c}{Within $t^\circ$ $\uparrow$} & \\
    \cmidrule(lr){6-7}\cmidrule(lr){8-10}
    & & & & & Mean & Median & 11.25 & 22.5 & 30 & \\
    \midrule
    \texttt{STL} & 38.30 & 63.76 & 0.6754 & 0.2780 & 25.01 & 19.21 & 30.14 & 57.20 & 69.15 & \\
    \midrule
    \texttt{LS} & 39.29 & 65.33 & 0.5493 & 0.2263 & 28.15 & 23.96 & 22.09 & 47.50 & 61.08 & 13.11 & 5.59  \\
    \texttt{SI} & 38.45 & 64.27 & 0.5354 & 0.2201 & 27.60 & 23.37 & 22.53 & 48.57 & 62.32 & 12.22 & 4.39 \\
    \texttt{RLW}~\citep{lin2021reasonable} & 37.17 & 63.77 & 0.5759 & 0.2410 & 28.27 & 24.18 & 22.26 & 47.05 & 60.62 & 15.78 & 7.78 \\
    \texttt{DWA}~\citep{liu2019end} & 39.11 & 65.31 & 0.5510 & 0.2285 & 27.61 & 23.18 & 24.17 & 50.18 & 62.39 & 11.89 & 3.57 \\
    \texttt{UW}~\citep{kendall2018multi} & 36.87 & 63.17 & 0.5446 & 0.2260 & 27.04 & 22.61 & 23.54 & 49.05 & 63.65 & 11.89 & 4.05 \\
    \texttt{MGDA}~\citep{desideri2012multiple} & 30.47 & 59.90 & 0.6070 & 0.2555 & 24.88 & 19.45 & 29.18 & \textbf{56.88} & \textbf{69.36} & 8.44 & 1.38 \\
    \texttt{PCGrad}~\citep{yu2020gradient} & 38.06 & 64.64 & 0.5550 & 0.2325 & 27.41 & 22.80 & 23.86 & 49.83 & 63.14 & 12.33 & 3.97 \\
    \texttt{GradDrop}~\citep{chen2020just} & 39.39 & 65.12 & 0.5455 & 0.2279 & 27.48 & 22.96 & 23.38 & 49.44 & 62.87 & 11.22 & 3.58 \\
    \texttt{CAGrad}~\citep{liu2021conflict} & 39.79 & 65.49 & 0.5486 & 0.2250 & 26.31 & 21.58 & 25.61 & 52.36 & 65.58 & 8.33 & 0.20 \\
    \texttt{IMTL-G}~\citep{liu2021towards} & 39.35 & 65.60 & 0.5426 & 0.2256 & 26.02 & 21.19 & 26.20 & 53.13 & 66.24 & 7.33 & -0.76 \\
    \texttt{MoCo}~\citep{fernando2022mitigating} & 40.30 & 66.07 & 0.5575 & 0.2135 & 26.67 & 21.83 & 25.61 & 51.78 & 64.85 & 7.78 & 0.16 \\
    \texttt{MoDo}~\citep{chen2023three} & 35.28 & 62.62 & 0.5821 & 0.2405 & 25.65 & 20.33 & 28.04 & 54.86 & 67.37 & 10.56 & 0.49 \\
    \texttt{Nash-MTL}~\citep{navon2022multi} & 40.13 & 65.93 & 0.5261 & 0.2171 & 25.26 & 20.08 & 28.40 & 55.47 & 68.15 & 4.33 & -4.04 \\
    \texttt{SDMGrad}~\citep{xiao2023direction} & 40.47 & 65.90 & 0.5225 & \textbf{0.2084} & 25.07 & 19.99 & 28.54 & 55.74 & 68.53 & \textbf{3.00} & \textbf{-4.84} \\
    \texttt{FAMO}~\citep{liu2024famo} & 38.88 & 64.90 & 0.5474 & 0.2194 & 25.06 & 19.57 & \textbf{29.21} & 56.61 & 68.98 & 5.11 & -4.10 \\
    \texttt{FAMO} (rerun)~\citep{liu2024famo} &38.60 & 63.29 & \textbf{0.5201} & 0.2185 & \textbf{24.50} & \textbf{19.36} & 27.48 &  53.75& 65.91& 6.11 & -3.27  \\
    \midrule
    \texttt{MARIGOLD} & \textbf{41.01}& \textbf{66.27} & 0.5270 & 0.2137  & 25.29  & 20.06 & 28.55 & 55.51 & 68.11  & 3.56 & -4.54\\
    \bottomrule
  \end{tabular}
  \end{adjustbox}
\end{table}

\begin{table}[!ht]
    \caption{Multi-task supervised learning on Cityscapes dataset.}
    \label{tab:cityscapes_app}
  \centering
    \begin{adjustbox}{max width=0.7\textwidth}
  \begin{tabular}{lllllll}
    \toprule
    \multirow{2}*{Method} & \multicolumn{2}{c}{Segmentation} & \multicolumn{2}{c}{Depth} & 
    \multirow{2}*{MR $\downarrow$} &
    \multirow{2}*{$\Delta k\%\downarrow$} \\
    \cmidrule(lr){2-3}\cmidrule(lr){4-5}
    & mIoU $\uparrow$ & Pix Acc $\uparrow$ & Abs Err $\downarrow$ & Rel Err $\downarrow$ & \\
    \midrule
    \texttt{STL} & 74.01 & 93.16 & 0.0125 & 27.77 & \\
    \midrule
    \texttt{LS} & 75.18 & 93.49 & 0.0155 & 46.77 & 10.25 & 22.60  \\
    \texttt{SI} & 70.95 & 91.73 & 0.0161 & 33.83 & 13.00 & 14.11 \\
    \texttt{RLW}~\citep{lin2021reasonable} & 74.57 & 93.41 & 0.0158 & 47.79 & 13.00 & 24.38 \\
    \texttt{DWA}~\citep{liu2019end} & 75.24 & 93.52 & 0.0160 & 44.37 & 10.00 & 21.45 \\
    \texttt{UW}~\citep{kendall2018multi} & 72.02 & 92.85 & 0.0140 & \textbf{30.13} & 9.00 & \textbf{5.89} \\
    \texttt{MGDA}~\citep{desideri2012multiple} & 68.84 & 91.54 & 0.0309 & 33.50 & 13.50 & 44.14  \\
    \texttt{PCGrad}~\citep{yu2020gradient} & 75.13 & 93.48 & 0.0154 & 42.07 & 10.25 & 18.29  \\
    \texttt{GradDrop}~\citep{chen2020just} & 75.27 & 93.53 & 0.0157 & 47.54 & 9.00 & 23.73  \\
    \texttt{CAGrad}~\citep{liu2021conflict} & 75.16 & 93.48 & 0.0141 & 37.60 & 8.75 & 11.64  \\
    \texttt{IMTL-G}~\citep{liu2021towards} & 75.33 & 93.49 & 0.0135 & 38.41 & 5.75 & 11.10 \\
    \texttt{MoCo}~\citep{fernando2022mitigating} & \textbf{75.42} & 93.55 & 0.0149 & 34.19 & 4.50 & 9.90 \\
    \texttt{MoDo}~\citep{chen2023three} & 74.55 & 93.32 & 0.0159 & 41.51 & 12.50 & 18.89 \\
    \texttt{Nash-MTL}~\citep{navon2022multi} & 75.41 & \textbf{93.66} & \textbf{0.0129} & 35.02 & \textbf{3.00} & 6.82 \\
    \texttt{SDMGrad}~\citep{xiao2023direction} & 74.53 & 93.52 & 0.0137 & 34.01 & 6.50 & 7.79 \\
    \texttt{FAMO}~\citep{liu2024famo} & 74.54 & 93.29 & 0.0145 & 32.59 & 9.00 & 8.13 \\
    \texttt{FAMO} (rerun)~\citep{liu2024famo} & 74.76 & 93.25 & 0.0140 & 35.78 & 9.75 & 9.92 \\
    \midrule
    \texttt{MARIGOLD} & 75.30 & 93.51 & 0.0139 & 34.79 & 5.25 & 8.62
 \\
    \bottomrule
  \end{tabular}
  \end{adjustbox}
\end{table}

In this section we provide more details of the implementation of our algorithms in the experiments. All experiments in Section~\ref{sec: public_exp} were conducted on a Tesla V100 GPU, and all experiments in Section~\ref{sec: industry_exp} were conducted in a distributed manner using 128 A100 GPUs.

For the lower-level problem model training part we follow the basic setup in \cite{liu2024famo, xiao2023direction} and set \texttt{Optim\_LL} to be \texttt{Adam} optimizer with learning rate equal to $10^{-4}$. The perturbation scale $r$ in Algorithm \ref{alg: hyper_marigold} is chosen as $10^{-3}$. To tackle the upper-level problem, we parameterize the weights vectors $\lambda$ and $\rho$ using softmax functions. To better illustrate the parameterization, we consider for simplicity the following optimization problem with log loss that achieves scale-invariant training.
\begin{align*}
    \min_{\theta}\sum_{i=1}^{m}w_i\log f_i(\theta)
\end{align*}
whose gradient can be written as
\begin{align*}
    \nabla_{\theta}\left(\sum_{i=1}^{m}w_i\log f_i(\theta)\right) = \sum_{i=1}^{m}\frac{w_i}{f_i(\theta)}\nabla f_i(\theta),
\end{align*}
which motivates the parameterization $\lambda = \softmax(\beta u / f),\ \rho = \softmax(\beta v / f)$ to guarantee the task weights are included in the probability simplex $\Delta^m$. In other words, in Algorithm \ref{alg: marigold} we have
\begin{align*}
    \lambda_i = \frac{e^{\beta u_i/f_i}}{{\sum_{j=1}^{m}e^{\beta u_j/f_j}}},\ \rho_i = \frac{e^{\beta v_i/f_i}}{{\sum_{j=1}^{m}e^{\beta v_j/f_j}}},
\end{align*}
where $u$ and $v$ are trainable parameters and we use \texttt{Adam} optimizer with learning rates both equal to $10^{-4}$. $\beta$ is the hyperparameter that are tuned on the set $\{0.01, 0.03, 0.1, 0.3, 1.0, 3.0\}$, and is chosen as $0.03$ and $1.0$ respectively for Cityscapes dataset and NYU-v2 dataset respectively. We also evaluate the loss balancing algorithms, whose numerical results can be found in Tables \ref{tab:nyuv2_loss} and \ref{tab:cityscapes_loss}. Furthermore, in Tables \ref{tab:nyuv2_app} and \ref{tab:cityscapes_app}, we combine both loss balancing and gradient balancing methods for a comprehensive comparison. As also observed in existing MTL works, loss balancing methods are usually faster than gradient balancing methods, as the former performs balancing based on loss function values with certain heuristics. In terms of the performance, the results in Tables \ref{tab:nyuv2_app} and \ref{tab:cityscapes_app} indicate that gradient balancing methods like \texttt{Nash-MTL} are usually better than loss balancing ones.

\bibliographystyle{plain}
\bibliography{bibfile}

\end{document}